\documentclass[conference]{IEEEtran}

\newcommand\blfootnote[1]{%
  \begingroup
  \renewcommand\thefootnote{}\footnote{#1}%
  \addtocounter{footnote}{-1}%
  \endgroup
}

\usepackage{times}
\usepackage[numbers]{natbib}
\usepackage{multicol}
\usepackage[bookmarks=true]{hyperref}

\usepackage{mathtools}

\usepackage{graphicx}
\usepackage{siunitx}
\usepackage{subcaption}
\usepackage{adjustbox}
\usepackage{lipsum}
\usepackage{comment}
\usepackage{float}

\title{DeepSocNav: Social navigation by imitating human behaviors}

\author{Juan Pablo de Vicente and Alvaro Soto$^{1}$
}

\begin{document}

\maketitle
\thispagestyle{empty}
\pagestyle{empty}

\blfootnote{*This work was partially funded by
FONDECYT grant 1181739}
\blfootnote{$^{1}$J.~de Vicente and A.~Soto, Dept.~of Computer Science, Pontificia 
Universidad Catolica de Chile,
        {\tt (jpdevicente@uc.cl, asoto@ing.puc.cl)}}%
\begin{abstract}

Current datasets to train social behaviors are usually borrowed from surveillance applications that capture visual data from a bird's-eye perspective. This leaves aside precious relationships and visual cues that could be captured through a first-person view of a scene. In this work, we propose a strategy to exploit the power of current game engines, such as Unity, to transform pre-existing bird's-eye view datasets into a first-person view, in particular, a depth view. Using this strategy, we are able to generate large volumes of synthetic data that can be used to pre-train a social navigation model. To test our ideas, we present DeepSocNav, a deep learning based model that takes advantage of the proposed approach to generate synthetic data. Furthermore, DeepSocNav includes a self-supervised strategy that is included as an auxiliary task. This consists of predicting the next depth frame that the agent will face. Our experiments show the benefits of the proposed model that is able to outperform relevant baselines in terms of social navigation scores.

\end{abstract}


\section{Introduction}


A desirable skill to operate in a human inhabited environment is to exercise social navigation, i.e., navigate in a way that does not hinder other humans present in the environment. Due to this, traditional navigation techniques based on algorithms such as A* or potential field are not applicable \cite{c47}. This is because they do not take into consideration people's intrinsic social behaviors.

Previous works have tried to solve this problem. In particular, previous attempts can be classified into two main groups: model-driven and data-driven approaches. On the one hand, model-driven approaches are characterized by an explicit use of relevant relationships and rules that guide navigational behaviors in a social context. Examples of model-driven approaches for social navigation are \cite{c1, c12, c14, c15, c16, c17, c23, c24, c26}, which mainly use the so-called social force \cite{c7} to model the social navigation of passers-by. As a major limitation, these approaches lack enough flexibility to handle unexpected or complex situations. 

On the other hand, data-driven approaches seek to learn social behaviors by extracting knowledge from a large number of examples. This has the advantage of automatically inferring complex relationships from data. Human behavior is highly stochastic \cite{c4,c9} and varies depending on factors such as culture \cite{c34}. As a consequence, the increased flexibility of data-driven techniques usually leads to models that outperform model-driven counterparts \cite{c3,c4,c5,c8,c30}. 

A relevant complexity to implement a data-driven approach for social navigation is the need for a large volume of data. Furthermore, the type of data collected plays a significant role in the quality of the final solution. In terms of social navigation, currently most public datasets consist of third-person recordings \cite{c49}, usually in bird's-eye view, with works that make use of this datasets such as \cite{c2,c6,c25,c30}. This means that models that directly use this data are limited to a 2D representation of the scene, which leaves aside useful relationships that can be captured through 3D observations and limits solutions to areas with a top side view or to use more expensive sensors such as LiDAR scanners to replicate a bird-eye view. A possible way to alleviate this problem is to record and tag first-person navigational data. Unfortunately, this approach does not scale properly, mainly due to the complexity of measuring the velocity and position of the first-person agent and the cost associated to capture a sizable amount of data.

This work helps to overcome this limitation by developing a virtual environment that allows us to transform the abundant bird's-eye view data to first-person views of all agents present in a giving scene. The simulated environment also features realistic human models and animations, increasing the information available to implement social navigational behaviors. 
Furthermore, it is also possible to include artificial agents to enrich the original data. Motions of these artificial agents do not ensure social behaviors but, as we show in this work, are useful to pre-train a model to acquire basic navigation skills.

In addition to the simulated environment, we also present DeepSocNav, a supervised learning model that seeks to imitate the social behaviors of real humans in crowded environments, using a first-person depth view. This model is capable of learning to replicate the navigational patterns of the agents present in the data. In this way, the model is encouraged to develop a social policy closer to what people do. Furthermore, to guide learning, DeepSocNav also learns the auxiliary task of predicting the next depth frame that the agent will face. This fosters the model to acquire a notion of space and agents in its visual field, thus improving its social performance and its ability to avoid collisions.


We support our contributions with suitable experiments to evaluate the main components of our model, conducting an ablation analysis to demonstrate its benefits.

\section{Proposed method}


\subsection{Description of the problem}
Current datasets to train social behaviors are usually borrowed from surveillance applications that capture visual data from a bird's-eye perspective \cite{c41}. This limits the scope of a model to consider only position and speed of passers-by. Furthermore, the resulting model is constrained to an operation where a bird's-eye camera is present or to rely on expensive equipment such as LiDAR sensors to simulate a bird's-eye view \cite{c30}. This situation motivates our proposed method that starting from a pedestrian's trajectory dataset corresponding to a bird's-eye perspective is able to recreate the corresponding first-person views, in particular, a view of depth information. This would enable a mobile robot to only require a depth camera to operate.


\subsection{First-person view data generation}\label{Sec:FirstPersonView}
For the recreation of the datasets using a first-person view, we develop a simulated environment using Unity, a video game development engine \cite{c48}. This environment loads previously designed maps and recreate the trajectory of agents using the data sequence of the agents' position in time. Specifically, the simulator moves each agent $i$ from its current position $(x_t^i, y_t^i)$ to next position $(x_{t+1}^i, y_{t+1}^i)$ at time instant $t+1$ until the agent reaches a target position $(G_x^i, G_y^i)$. These displacements follow a realistic walking animation through Unity's animation control asset \cite{c35}.

\begin{figure*}
  \begin{subfigure}{0.32\textwidth}
  \vspace*{0.1in}
    \center{\includegraphics[width=0.8\textwidth]{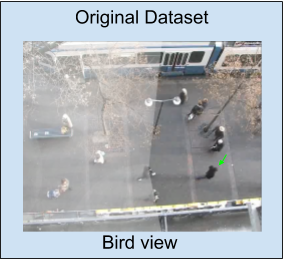}}
    \caption{BIWI dataset: hotel top-down frame.} \label{fig:1a}
  \end{subfigure}%
  \hspace*{\fill}   
  \begin{subfigure}{0.634\textwidth}
  \vspace*{0.1in}
    \center{\includegraphics[width=0.8\textwidth]{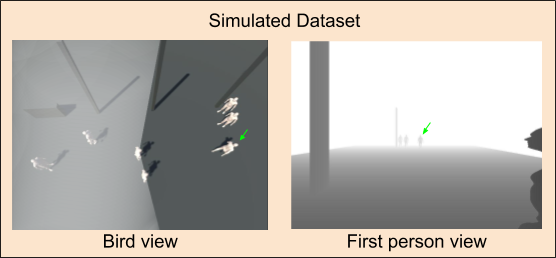}}
    \caption{Simulated top-down and first-person views.} \label{fig:1b}
  \end{subfigure}%
\caption{a) Representative frame from the real scene present in the original dataset. b) Top and first person depth views of the frame in a) recreated using the simulator.} \label{fig:1}
\end{figure*}

Using the simulation, for each agent and time instant $t$, a first-person depth view is stored, together with its current coordinate $(x_t^i, y_t^i)$, target coordinate $(G_x^i, G_y^i)$, and velocity $(vx_t^i, vy_t^i)$. From this data, it is expected that the model can learn the implicit social behaviors underlying the trajectories of the agents in the original dataset. An example of a recreated frame in the simulated environment can be seen in Figure \ref{fig:1}. Specifically, Figure 1a corresponds to a frame in the original video and Figure 1b to its re-creation in the simulated environment using a first-person depth view.


\subsection{Data Augmentation}\label{Sec:DataAugmentation}
Besides its bird's-eye view, a second limitation of current pedestrian trajectory datasets is the limited size. 
Purely pedestrian datasets have usually a small volume, in the order of 1K to 15K people in total \cite{c41}. Furthermore, some of these datasets do not include rich interactions among people, where most pedestrian follow straight trajectories to reach their goals. This may produce models that overlook the need to learn social navigation skills, such as turning and deviating from a straight line in order to avoid collisions.

To alleviate the data scarcity problem, we propose a data augmentation scheme that takes advantage of the autonomous navigation capabilities provided by the library NavMesh included in Unity \cite{c35}. Specifically, using Unity, we can add virtual agents to a scene that are instructed to reach randomly generated goal positions. Using this scheme, we can generate an almost unlimited amount of navigational trajectories. As a main limitation, this artificial trajectories do not ensure social behaviors. However, we can use them as a massive source of data to train an initial navigational model that is able to reach target positions while avoiding obstacles. Afterwards, we can fine-tune this model using synthetic data that incorporates social behaviors. 

\subsection{DeepSocNav}

Following the usual setup \cite{c6,c7}, social navigation is seen as a problem of velocity prediction given a sequence of observations. For each time instant $t$ and agent $i$, we have a measurement of its current position $(x_t^i, y_t^i)$, the coordinates of its goal position $G=(Gx_t^i, Gy_t^i)$, and a depth image $D_t^i$ corresponding to what the agent sees at time $t$ according to a first-person view. From a sequence of observations $t-T$ to $t$, the agent velocity $(vx_{t+1}^i, vy_{t+1}^i)$ is predicted for the following time step $t+1$. In this way the model can be defined as a function $f({s_{t-T},...,s_t}, G)$ described as follows:

\vspace{-0.3cm}
\begin{equation}
\begin{aligned}
\label{eq:model_equation}
f({s_{t-T},...,s_t}, G) =\Vec{v}_{t+1}, \;\; \;\; s_t = (x_t, y_t, D_t) 
\end{aligned}
\end{equation}

\begin{figure*}[!htb]
    \center{\includegraphics[width=0.8\textwidth]
    {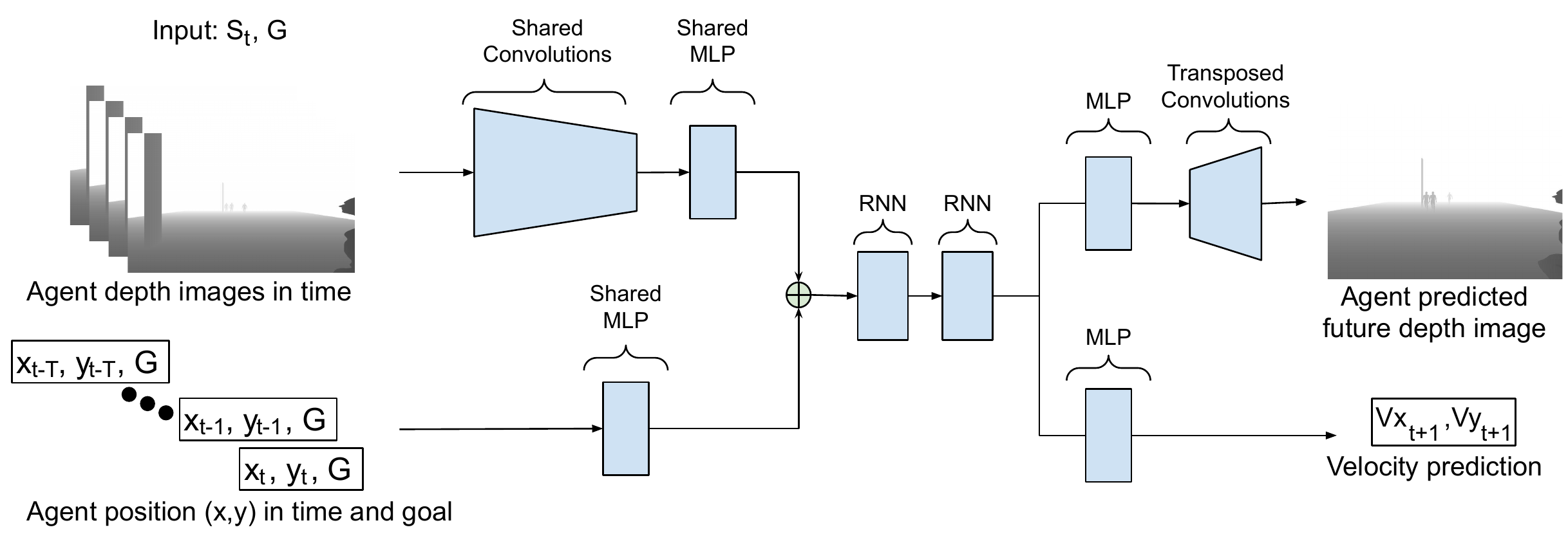}}
    \caption{\label{fig:my-label} DeepSocNav overall architecture, it consists of two heads: i) Head 1 predicts velocity of an agent at time $t+1$ given a history of previous observations from steps $t-T$ to $t$, ii) Head 2 predicts depth image $\hat D_{t+1}$ for next time step $t+1$.}
    \label{fig:overall_architecture}
\end{figure*}

The positional entries $(x_t,y_t)$ and target $(Gx, Gy)$ are normalized from the largest and smallest value possible among all the maps. Because depth cameras have a limited range, we imitate this limitation by defining the pixel value $p_{\bf{x,y},t}$ in depth images as follows:  

\[
    p_{\bf{x,y},t}= 
\tag{2}
\begin{dcases}
    d_{\bf{x,y},t}/d_{max},& \text{if } d_{\bf{x,y},t} < d_{max} \\
    1,              & \text{otherwise}
\end{dcases}
\]

\noindent where $d_{\bf{x,y}}$ corresponds to the distance between the agent's camera and the surface located on the pixel $p_{\bf{x,y}}$; $d_{max}$ is the maximum range distance provided by the depth camera.

Figure~\ref{fig:overall_architecture} shows a schematic view of the learning architecture behind DeepSocNav. Each depth frame of the history of T observations is processed by a shared convolutional network followed by a MLP. Also, an embedding of the current position and goal of the agent is obtained with a shared MLP. These embeddings are concatenated to feed a 2 layer LSTM. Afterwards, the hidden state of the last cell of the LSTM is used to feed two prediction heads that are explained next.


\subsubsection{Head 1: Velocity prediction}

The goal of this head is to predict the velocity $\Vec{v}_{t+1}$ that the agent must take given observations $s_t$ corresponding to the last $T$ instants of time, so that the agent arrives at the destination in a socially correct way. For this, the output of the LSTM is processed by an MLP, where the last one outputs $(vx_{t+1}^i, vy_t^i)$.

\subsubsection{Head 2: Forecasting the future as an auxiliary task}
The goal of this head is to predict the future depth image $\hat D_{t+1}$. This is added as an auxiliary task, which seeks to guide learning by encouraging the model to learn the features needed to predict how the scene will look in the close future. Our hypothesis is that the ability to anticipate where objects and agents in a scene will move is closely related to the ability to know how they act.

\subsubsection{Loss function}

Following the two heads, the resulting loss function $L = L_v + k L_D$ consists of two terms that predict velocity and depth information, respectively, where $k$ is a weighting constant. These terms are defined as follow: 
\vspace{-0.3cm}
\begin{equation*}
\tag{3}
\begin{aligned}
L_v = \sum \|v_{t+1} - \hat{v}_{t+1}\|^2w(t) \\ 
L_D = \sum \|D_{t+1} - \hat{D}_{t+1}\|^2w(t)
\end{aligned}
\end{equation*}

\[
    w(t)= 
\tag{4}
\begin{dcases}
    c,& \text{if } min(p_{\bf{x,y},t}) < \beta \\
    1,              & \text{otherwise}
\end{dcases}
\]

To prioritize learning behaviors that involve close obstacles, we introduce the weighting $w(t)$ in (4) which assigns a weight $c$, where $c>1$, to the output of the model when the depth camera detects an object at a distance less than $\beta$.

\section{Experiments}

To test our model, we use as a benchmark the ETH BIWI Walking Pedestrians dataset (BIWI) \cite{c16}. This dataset corresponds to RGB images from a bird's-eye view of two different scenarios: ETH and Hotel, for a total of 787 agents. We recreate the scenes of this datasets in Unity using our proposed methodology, extracting depth images using a first-person view of each trajectory. We use the resulting data to train our proposed model: DeepSocNav. Furthermore, using the data augmentation scheme described in Section \ref{Sec:DataAugmentation}, we generate a total of 6,000 artificial trajectories on both maps. 




In our experiments, we generate depth images using a resolution of 320x240 pixels, a field of view of \ang{135}, and maximum range of $d_{max} = 7[m]$. To train DeepSocNav, both synthetic and BIWI trajectories are randomly separated in $\frac{9}{10}$ for training and $\frac{1}{10}$ evaluation. In our experiments, we use $T=10$. Also, we use $c=2$ to weight the predictions when an obstacle is nearby, and $k=0.1$ in the  loss function.

As we describe in Section \ref{Sec:DataAugmentation}, we use the synthetic trajectories to pre-train DeepSocNav. Specifically, we pre-train for 15 epochs, using a learning rate of 0.001 and Adam as the optimizer. Afterwards, DeepSocNav is fine-tuned for 10 epochs using the synthetic depth images corresponding to the BIWI dataset, Adam as the optimizer, and a learning rate of 0.0001. During testing, we use DeepSocNav to control the navigation policy of the target agent at 10 [Hz]. Goal positions follow the BIWI dataset, denoted as a circular area with a radius of $1.5[mts]$. In average, trajectories consist of $9[mts]$ length.


\vspace{0.1cm}

\noindent \textbf{Baselines.} We consider the following baselines:


\begin{itemize}
    \item \textbf{Reciprocal Velocity Obstacles (RVO)}. RVO is implemented in NavMesh, the navigational system included in Unity \cite{c42}. 
    \item \textbf{Social Force Model (SFM)}. In our implementation, we use the hyperparameters described in \cite{c43}. Furthermore, we only consider the forces of agents and obstacles present in the visual range of the target agent. 
    \item \textbf{NaviGAN}. NaviGAN is a generative navigation algorithm that uses generative adversarial networks (GANs), to learn how to generate routes that seek to optimize the comfort and naturalness of these routes \cite{c45}. 
\end{itemize}

\vspace{0.1cm}

\noindent \textbf{DeepSocNav ablation study.}
To study the contribution of the different parts of the model, the following variations of DeepSocNav are also considered.
\begin{itemize}
    \item $DeepSocNav_{noAux}$: Does not include the auxiliary task of predicting the depth image for the next frame.
    \item $DeepSocNav_{T1}$: Uses  $T = 1$ as the time window.
    \item $DeepSocNav_{halfPreTrain}$: Only uses half of the trajectories for pre-train.
    \item $DeepSocNav_{noPreTrain}$: Does not include pre-training.
\end{itemize}

\begin{table}[H]
\centering
\begin{tabular}{|l|l|l|l|}
\hline
Model               & Social Score   & Collisions     & Success    \\ \hline
GT                  & 0.034          & -              & -          \\ \hline
RVO                 & 0.067          & -              & \textbf{1} \\ \hline
SFM                 & 0.054          & 0.037          & \textbf{1} \\ \hline
DeepSocNav          & \textbf{0.040} & \textbf{0.018} & \textbf{1} \\ \hline
\end{tabular}
\caption{Social results in online evaluation}
\label{table:Social_Results}
\end{table}


\begin{table}[H]
\centering
\begin{tabular}{|l|l|l|l|l|}
\hline
Model               & ADE [m]        & FDE [m]         \\ \hline
RVO                 & \textbf{0.20}          & 0.20    \\ \hline
SFM                 & \textbf{0.20} & \textbf{0.16}    \\ \hline
NaviGAN \cite{c44}  & 0.43          & 0.74             \\ \hline
DeepSocNav          & 0.24          & 0.34             \\ \hline
\end{tabular}
\caption{Distance relative to GT during testing.}
\label{table:Distance_results}
\end{table}


\begin{table}[H]
\centering
\vspace{0.1in}
\begin{tabular}{|l|l|l|l|l|l|}
\hline
Model & Social& Coll-. & ADE  & FDE  \\
 & Score & ission  & [m] & [m] \\
\hline
DeepSocNav & \textbf{0.040} & \textbf{0.018} & \textbf{0.24} & 0.34 \\ \hline
$DeepSocNav_{noAux}$ & 0.042 & 0.037 & \textbf{0.24} & \textbf{0.20} \\ \hline
$DeepSocNav_{T1}$ & 0.051 & \textbf{0.018} & 0.32 & 0.52 \\ \hline
$DeepSocNav_{halfPreTrain}$ & 0.041 & 0.056 & 0.27 & 0.34 \\ \hline
$DeepSocNav_{noPreTrain}$ & 0.047 & 0.094 & 0.33 & 0.41 \\ \hline
\end{tabular}
\caption{Ablation results of online evaluation.}
\label{table:Sensibility_results}
\end{table}

\vspace{-0.3cm}

\noindent \textbf{Metrics.} The following metrics are used for evaluation.  
\begin{itemize}
    \item Social Score: Penalizes the agent with a cost $c_l$ each time that it is inside a personal circle of another agent. We define 3 circles of radii $r_1=0.5[m]$, $r_2=0.75[m]$, $r_3=1.0[m]$ with a cost of $c_1=1$, $c_2=0.5$ and $c_3=0.1$ respectively.
    \item Average Distance Error (ADE) with respect to GT.
    \item Final Distance Error (FDE) with respect to goal.
    \item Collisions: number of collisions.
    \item Success: successful goal reaching.
\end{itemize}

\section{Results and Analysis}

Table \ref{table:Social_Results} and \ref{table:Distance_results} shows the performance of DeepSocNav and the baselines. In terms of the social scores (Social Score and Collisions), DeepSocNav outperforms all the baselines. In terms of success score, DeepSocNav has a perfect score. In terms of the distance metrics (ADE and FDE), while DeepSocNav has a competitive performance, SFM is able to generate the most similar trajectories to the GT. However, it is important to note that SFM uses explicit information about the positions and velocities of the other agents, while DeepSocNav only uses a first-person depth view.  

We believe that by taking advantage of its first-person depth view, DeepSocNav is able to learn and then infer rich information about the intentions and potential trajectories of other passers-by. This explains its advantages in the performance with respect to state-of-the art techniques such as NaviGAN \cite{c45}, which corresponds to a deep learning model that uses oracle information from other passers-by at the coordinate level. 

\subsection{Ablation analysis}
Table \ref{table:Sensibility_results} shows an ablation analysis with respect to the main components behind DeepSocNav. It is possible to note that although the full model achieves the best results all around, $DeepSocNav_{noAux}$ ties it on ADE and scores the best result in FDE. This means that predicting depth information for the next frame helps to avoid collisions and to improve social navigation, but there is a trade-off in terms of generating trajectories that are similar to the GT over longer distances.

The version of DeepSocNav with the worst performance overall is $DeepSocNav_{T1}$. This supports the relevance of keeping a small memory to perform correctly in the task. However, by using the auxiliary task, the models is still able to avoid collisions.

Finally, $DeepSocNav_{noPreTrain}$ is the variant with the highest number of collisions. This supports our initial hypothesis about the relevance of pre-training the model using simulated data to avoid data scarcity problems. In particular, our pre-training strategy generates an initial navigation scheme with obstacle avoidance and goal-reaching capabilities. This can also be seen by considering our full model and $DeepSocNav_{halfPreTrain}$, where $DeepSocNav_{halfPreTrain}$ has almost the same Social Score than the full model, but more than double the collisions.

\section{Conclusions}
In this work, we show the advantage of using the power of current game engines, such as Unity, to improve the generation of data to train social navigation agents. In particular, we show the advantage of exploiting a method to extract first-person view recordings of passers-by from bird-view datasets. Similarly, we show the advantage of using simulated data from a virtual environment to pre-train a model to obtain an initial navigation scheme that include obstacle avoidance and goal-reaching capabilities.

In particular, we demonstrate the advantages of the information-rich first-person view by training DeepSocNav, an LSTM-based model capable of navigating through a crowded environment. Our results validate the impact of our proposed strategy in terms of social navigation by outperforming all baselines considered in this work. We also demonstrate the relevance of using a short-term memory of previous views, as well as, a prediction of the next depth frame, which is included as an auxiliary task. We believe that the use of this type of auxiliary tasks and self-supervised learning strategies might play an important role to improve current social navigation models.   







\newpage
\bibliography{IEEEabrv,mybibfile}

\begin{thebibliography}{29}
\providecommand{\natexlab}[1]{#1}
\providecommand{\url}[1]{\texttt{#1}}
\expandafter\ifx\csname urlstyle\endcsname\relax
  \providecommand{\doi}[1]{doi: #1}\else
  \providecommand{\doi}{doi: \begingroup \urlstyle{rm}\Url}\fi

\bibitem[c34(2011)]{c34}
The wisdom of crowds, December 2011.
\newblock URL
  \url{https://www.economist.com/christmas-specials/2011/12/17/the-wisdom-of-crowds}.

\bibitem[{Alahi} et~al.(2014){Alahi}, {Ramanathan}, and {Fei-Fei}]{c1}
A.~{Alahi}, V.~{Ramanathan}, and L.~{Fei-Fei}.
\newblock Socially-aware large-scale crowd forecasting.
\newblock In \emph{2014 IEEE Conference on Computer Vision and Pattern
  Recognition}, pages 2211--2218, 2014.

\bibitem[{Alahi} et~al.(2016){Alahi}, {Goel}, {Ramanathan}, {Robicquet},
  {Fei-Fei}, and {Savarese}]{c2}
A.~{Alahi}, K.~{Goel}, V.~{Ramanathan}, A.~{Robicquet}, L.~{Fei-Fei}, and
  S.~{Savarese}.
\newblock Social lstm: Human trajectory prediction in crowded spaces.
\newblock In \emph{2016 IEEE Conference on Computer Vision and Pattern
  Recognition (CVPR)}, pages 961--971, 2016.

\bibitem[Amirian et~al.(2020)Amirian, Zhang, Castro, Baldelomar, Hayet, and
  Pettre]{c49}
Javad Amirian, Bingqing Zhang, Francisco~Valente Castro, Juan~Jose Baldelomar,
  Jean-Bernard Hayet, and Julien Pettre.
\newblock Opentraj: Assessing prediction complexity in human trajectories
  datasets.
\newblock In \emph{Asian Conference on Computer Vision (ACCV)}, number CONF.
  Springer, 2020.

\bibitem[{Chen} et~al.(2017){Chen}, {Everett}, {Liu}, and {How}]{c4}
Y.~F. {Chen}, M.~{Everett}, M.~{Liu}, and J.~P. {How}.
\newblock Socially aware motion planning with deep reinforcement learning.
\newblock In \emph{2017 IEEE/RSJ International Conference on Intelligent Robots
  and Systems (IROS)}, pages 1343--1350, 2017.

\bibitem[Chen et~al.(2016)Chen, Liu, Everett, and How]{c3}
Yu~Fan Chen, Miao Liu, Michael Everett, and Jonathan~P. How.
\newblock Decentralized non-communicating multiagent collision avoidance with
  deep reinforcement learning.
\newblock \emph{CoRR}, abs/1609.07845, 2016.
\newblock URL \url{http://arxiv.org/abs/1609.07845}.

\bibitem[{Fahad} et~al.(2018){Fahad}, {Chen}, and {Guo}]{c5}
M.~{Fahad}, Z.~{Chen}, and Y.~{Guo}.
\newblock Learning how pedestrians navigate: A deep inverse reinforcement
  learning approach.
\newblock In \emph{2018 IEEE/RSJ International Conference on Intelligent Robots
  and Systems (IROS)}, pages 819--826, 2018.

\bibitem[Ferrer et~al.(2016)Ferrer, Zulueta, Cotarelo, and Sanfeliu]{c43}
Gonzalo Ferrer, Anaís Zulueta, Fernando Cotarelo, and A.~Sanfeliu.
\newblock Robot social-aware navigation framework to accompany people walking
  side-by-side.
\newblock \emph{Autonomous Robots}, 41, July 2016.
\newblock \doi{10.1007/s10514-016-9584-y}.

\bibitem[Gupta et~al.(2018)Gupta, Johnson, Fei{-}Fei, Savarese, and Alahi]{c6}
Agrim Gupta, Justin Johnson, Li~Fei{-}Fei, Silvio Savarese, and Alexandre
  Alahi.
\newblock Social {GAN:} socially acceptable trajectories with generative
  adversarial networks.
\newblock \emph{CoRR}, abs/1803.10892, 2018.
\newblock URL \url{http://arxiv.org/abs/1803.10892}.

\bibitem[{Hamandi} et~al.(2019){Hamandi}, {D’Arcy}, and {Fazli}]{c30}
M.~{Hamandi}, M.~{D’Arcy}, and P.~{Fazli}.
\newblock Deepmotion: Learning to navigate like humans.
\newblock In \emph{2019 28th IEEE International Conference on Robot and Human
  Interactive Communication (RO-MAN)}, pages 1--7, 2019.

\bibitem[Helbing and Molnar(1998)]{c7}
Dirk Helbing and Peter Molnar.
\newblock Social force model for pedestrian dynamics.
\newblock \emph{Physical Review E}, 51, May 1998.
\newblock \doi{10.1103/PhysRevE.51.4282}.

\bibitem[Joo et~al.(2019)Joo, Simon, Cikara, and Sheikh]{c8}
Hanbyul Joo, Tomas Simon, Mina Cikara, and Yaser Sheikh.
\newblock Towards social artificial intelligence: Nonverbal social signal
  prediction in {A} triadic interaction.
\newblock \emph{CoRR}, abs/1906.04158, 2019.
\newblock URL \url{http://arxiv.org/abs/1906.04158}.

\bibitem[Kretzschmar et~al.(2016)Kretzschmar, Spies, Sprunk, and Burgard]{c9}
Henrik Kretzschmar, Markus Spies, C.~Sprunk, and W.~Burgard.
\newblock Socially compliant mobile robot navigation via inverse reinforcement
  learning.
\newblock \emph{The International Journal of Robotics Research}, 35:\penalty0
  1289 -- 1307, 2016.

\bibitem[{Leal-Taixé} et~al.(2011){Leal-Taixé}, {Pons-Moll}, and
  {Rosenhahn}]{c12}
L.~{Leal-Taixé}, G.~{Pons-Moll}, and B.~{Rosenhahn}.
\newblock Everybody needs somebody: Modeling social and grouping behavior on a
  linear programming multiple people tracker.
\newblock In \emph{2011 IEEE International Conference on Computer Vision
  Workshops (ICCV Workshops)}, pages 120--127, 2011.

\bibitem[Lerner et~al.(2007)Lerner, Chrysanthou, and Lischinski]{c14}
Alon Lerner, Yiorgos Chrysanthou, and Dani Lischinski.
\newblock Crowds by example.
\newblock \emph{Comput. Graph. Forum}, 26:\penalty0 655--664, September 2007.
\newblock \doi{10.1111/j.1467-8659.2007.01089.x}.

\bibitem[{Luber} et~al.(2010){Luber}, {Stork}, {Tipaldi}, and {Arras}]{c15}
M.~{Luber}, J.~A. {Stork}, G.~D. {Tipaldi}, and K.~O. {Arras}.
\newblock People tracking with human motion predictions from social forces.
\newblock In \emph{2010 IEEE International Conference on Robotics and
  Automation}, pages 464--469, 2010.

\bibitem[{Pellegrini} et~al.(2009){Pellegrini}, {Ess}, {Schindler}, and {van
  Gool}]{c16}
S.~{Pellegrini}, A.~{Ess}, K.~{Schindler}, and L.~{van Gool}.
\newblock You'll never walk alone: Modeling social behavior for multi-target
  tracking.
\newblock In \emph{2009 IEEE 12th International Conference on Computer Vision},
  pages 261--268, 2009.

\bibitem[Pellegrini et~al.(2010)Pellegrini, Ess, and Van~Gool]{c17}
Stefano Pellegrini, Andreas Ess, and Luc Van~Gool.
\newblock Improving data association by joint modeling of pedestrian
  trajectories and groupings.
\newblock In Kostas Daniilidis, Petros Maragos, and Nikos Paragios, editors,
  \emph{Computer Vision -- ECCV 2010}, pages 452--465, Berlin, Heidelberg,
  2010. Springer Berlin Heidelberg.
\newblock ISBN 978-3-642-15549-9.

\bibitem[Robicquet et~al.(2016)Robicquet, Sadeghian, Alahi, and Savarese]{c41}
Alexandre Robicquet, Amir Sadeghian, Alexandre Alahi, and Silvio Savarese.
\newblock Learning social etiquette: Human trajectory understanding in crowded
  scenes.
\newblock In \emph{Computer Vision -- ECCV 2016}, volume 9912, pages 549--565,
  October 2016.
\newblock ISBN 978-3-319-46483-1.
\newblock \doi{10.1007/978-3-319-46484-8_33}.

\bibitem[Siegwart et~al.(2004)Siegwart, Nourbakhsh, and Scaramuzza]{c47}
R.~Siegwart, I.~Nourbakhsh, and D.~Scaramuzza.
\newblock Introduction to autonomous mobile robots.
\newblock 2004.

\bibitem[Technologies(2018)]{c35}
Unity Technologies.
\newblock \emph{Unity User Manual}, 2018.
\newblock URL \url{https://docs.unity3d.com/Manual/index.html}.

\bibitem[{Trautman} and {Krause}(2010)]{c23}
P.~{Trautman} and A.~{Krause}.
\newblock Unfreezing the robot: Navigation in dense, interacting crowds.
\newblock In \emph{2010 IEEE/RSJ International Conference on Intelligent Robots
  and Systems}, pages 797--803, 2010.

\bibitem[Treuille et~al.(2006)Treuille, Cooper, and Popovi\'{c}]{c24}
Adrien Treuille, Seth Cooper, and Zoran Popovi\'{c}.
\newblock Continuum crowds.
\newblock \emph{ACM Trans. Graph.}, 25\penalty0 (3):\penalty0 1160–1168, July
  2006.
\newblock ISSN 0730-0301.
\newblock \doi{10.1145/1141911.1142008}.
\newblock URL \url{https://doi.org/10.1145/1141911.1142008}.

\bibitem[Tsai(2019)]{c44}
Chieh-En Tsai.
\newblock A generative approach for socially compliant navigation.
\newblock Master's thesis, Pittsburgh, PA, June 2019.

\bibitem[Tsai and Oh(2020)]{c45}
Chieh-En Tsai and Jean Oh.
\newblock Navigan: A generative approach for socially compliant navigation,
  2020.

\bibitem[{Unity Technologies}(2018)]{c48}
{Unity Technologies}.
\newblock Unity, May 2018.
\newblock URL \url{https://unity.com}.
\newblock 2018.1.2f1.

\bibitem[{van den Berg} et~al.(2008){van den Berg}, {Ming Lin}, and
  {Manocha}]{c42}
J.~{van den Berg}, {Ming Lin}, and D.~{Manocha}.
\newblock Reciprocal velocity obstacles for real-time multi-agent navigation.
\newblock In \emph{2008 IEEE International Conference on Robotics and
  Automation}, pages 1928--1935, 2008.

\bibitem[Vemula et~al.(2017)Vemula, M{\"{u}}lling, and Oh]{c25}
Anirudh Vemula, Katharina M{\"{u}}lling, and Jean Oh.
\newblock Social attention: Modeling attention in human crowds.
\newblock \emph{CoRR}, abs/1710.04689, 2017.
\newblock URL \url{http://arxiv.org/abs/1710.04689}.

\bibitem[{Yamaguchi} et~al.(2011){Yamaguchi}, {Berg}, {Ortiz}, and {Berg}]{c26}
K.~{Yamaguchi}, A.~C. {Berg}, L.~E. {Ortiz}, and T.~L. {Berg}.
\newblock Who are you with and where are you going?
\newblock In \emph{CVPR 2011}, pages 1345--1352, 2011.

\end{thebibliography}



\end{document}